\documentclass[manuscript]{acmart}

\usepackage{amsmath}

\DeclareMathOperator*{\argmin}{arg\,min}

\usepackage{graphicx}
\usepackage{subfigure}

\AtBeginDocument{%
  \providecommand\BibTeX{{%
    \normalfont B\kern-0.5em{\scshape i\kern-0.25em b}\kern-0.8em\TeX}}}

\copyrightyear{2021} 
\acmYear{2021}  
\setcopyright{acmcopyright}\acmConference[IUI '21]{26th International Conference on Intelligent User Interfaces}{April     14--17, 2021}{College Station, TX, USA}
\acmBooktitle{26th International Conference on Intelligent User Interfaces (IUI '21), April     14--17, 2021, College Station, TX, USA}
\acmPrice{15.00}
\acmDOI{10.1145/3397481.3450670}
\acmISBN{978-1-4503-8017-1/21/04}

\begin{document}

\title{DeepSI: Interactive Deep Learning for Semantic Interaction}


\author{Yali Bian}
\email{yali@vt.edu}
\affiliation{%
 \institution{Virginia Tech}
 \city{Blacksburg}
 \state{Virginia}
 \country{United States}}

\author{Chris North}
\email{north@vt.edu}
\affiliation{%
 \institution{Virginia Tech}
 \city{Blacksburg}
 \state{Virginia}
 \country{United States}}


\renewcommand{\shortauthors}{Bian and North}

\begin{abstract}

In this paper, we design novel interactive deep learning methods to improve semantic interactions in visual analytics applications.
The ability of semantic interaction to infer analysts' precise intents during sensemaking is dependent on the quality of the underlying data representation.
We propose the $\text{DeepSI}_{\text{finetune}}$ framework that integrates deep learning into the human-in-the-loop interactive sensemaking pipeline, with two important properties.
First, deep learning extracts meaningful representations from raw data, which improves semantic interaction inference. 
Second, semantic interactions are exploited to fine-tune the deep learning representations, which then further improves semantic interaction inference. This feedback loop between human interaction and deep learning enables efficient learning of user- and task-specific representations. 
To evaluate the advantage of embedding the deep learning within the semantic interaction loop, we compare $\text{DeepSI}_{\text{finetune}}$ against a state-of-the-art but more basic use of deep learning as only a feature extractor pre-processed outside of the interactive loop.
Results of two complementary studies, a human-centered qualitative case study and an algorithm-centered simulation-based quantitative experiment, show that $\text{DeepSI}_{\text{finetune}}$ more accurately captures users' complex mental models with fewer interactions.

\end{abstract}



\begin{teaserfigure}
  \includegraphics[width=\textwidth]{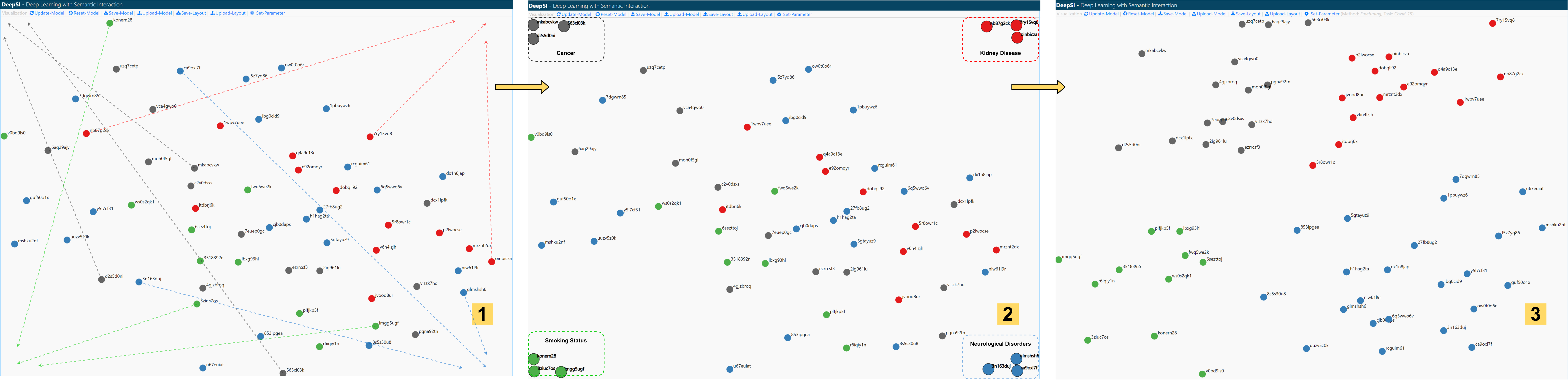}
  \caption{Screenshots during the analysis of COVID-19 research articles about four risk factors~(depicted in different colors) using our proposed model $\text{DeepSI}_{\text{finetune}}$: 
  (1) the initial layout of all articles projected from pretrained BERT representations of the raw text data;
  (2) the analyst performs semantic interactions to provide visual feedback regarding articles about different risk factors; these interactions are then exploited to tune the underlying DL model BERT;
  (3) the resulting projection updated by the tuned BERT.}
  \label{fig:covid19-1}
\end{teaserfigure}
\maketitle

\section{Introduction}
\label{sec:intro}

Semantic interaction~(SI)~\cite{7160906,endert2012semantic} is an interaction methodology that is commonly utilized to enhance visual analytics~(VA) systems.
SI-enabled systems let the analyst   directly manipulate interactive projections of data~\cite{Self:2016:BGU:2939502.2939505}.
The semantic meaning behind these projection interactions is the similarity relationships the analyst wishes to find within the data during the sensemaking process~\cite{pirolli_2005}.
As shown in Fig.~\ref{fig:covid19-1}-2, the analyst  drags 12 COVID-19 article points into four clusters to provide the visual feedback of grouping articles based on their perceived relevant risk factors.
With these intuitive and natural interactions, the analyst can remain within the cognitive zone~\cite{green2009building}, thereby enhancing the analyst's efficiency in performing analytic tasks~\cite{10.1145/3377325.3377516}. 
In the system, an interactive dimensionality reduction~(DR) component~\cite{10.1145/3377325.3377516,sacha2016visual} plays a key role in capturing the analyst's intent behind these interactions by learning a new projection layout~(Fig.~\ref{fig:covid19-1}-3).
To determine the analyst's precise intent, increasingly powerful interactive DR models~\cite{10.1145/3377325.3377516} have been proposed, from linear~\cite{Leman:2013it,6634115,House:2015hs} to non-linear models~\cite{10.1145/3311790.3396646,7534876}, and from single-model to multi-model approaches~\cite{dowling2018bidirectional,Bradel:dd,10.1016/j.bdr.2019.04.003,wenskovitch2017observation}.

However, the ability of semantic interaction to infer analysts’ precise intents during sensemaking is dependent on the quality of the underlying data representation.
Deep learning~(DL)~\cite{LeCun:2015dt} is a state-of-the-art representation learning method~\cite{10.1109/tpami.2013.50}, which can automatically extract abstract and useful hierarchical representations from raw data~\cite{6472238}.
This offers the new opportunity to power SI in capturing the analyst's intent. 
We denote the DL-enhanced SI system as \textbf{DeepSI}.
Previous researches have shown that even the usage of the pretrained DL representations as fixed data features have better performance than hand-crafted features in SI-enabled VA systems~\cite{bian2019deepva,bian2019evaluating}. 
We denote this straightforward DeepSI design with the basic use the pretrained DL as only a feature extractor in SI pipeline as $\text{DeepSI}_{\text{vanilla}}$.

In this paper, we aim to further improve semantic interaction inference by fine-tuning the model to obtain user- and task-specific representations from the pretrained DL model.
Central to this design goal are two research questions: 
\begin{itemize}
    \item \textit{How to exploit semantic interactions to \underline{accurately} adapt the pretrained representations to current analytic tasks?}  

    \item \textit{How to make \underline{efficient} adaptations, so that a small number of semantic interactions are enough for analysts to express their intents?} 
\end{itemize}

To address these two questions, we propose a novel DeepSI framework, $\text{DeepSI}_{\text{finetune}}$, with the following two design goals. 
First, we 
insert the interactive DL training into the bidirectional structure of the semantic interaction pipeline, so that  interactions trigger the DL adaptation.
Thereby, new user- and task-specific representations are generated  based on semantic interactions provided by analysts during their sensemaking process.
Second, we employ the fine-tuning based DL adaption approach and the MDS-based interactive DR model to minimize the number of parameters that require training in the underlying model.
Therefore, $\text{DeepSI}_{\text{finetune}}$ can tune the DL model efficiently from the analyst's interactions without information loss.
Specifically, we use the pretrained BERT~\cite{devlin2018bert}, a state-of-the-art DL model for NLP tasks, as the DL model representative inside $\text{DeepSI}_{\text{finetune}}$ for visual text analysis tasks.

To assess how well $\text{DeepSI}_{\text{finetune}}$ addresses these questions by integrating DL into the semantic interaction loop, we compare it with the well-evaluated baseline model $\text{DeepSI}_{\text{vanilla}}$~\cite{bian2019deepva,bian2019evaluating}, which uses DL outside of the interactive loop, in two complementary experiments:
a human-centered qualitative case study about COVID-19 academic articles; and
an algorithm-centered simulation-based quantitative analysis of three commonly used text corpora: Standford Sentiment Treebank~(SST), Vispubdata, and 20 Newsgroups. 
The results of both experiments show that $\text{DeepSI}_{\text{finetune}}$ not only captures the analyst's precise intent more accurately, but also requires fewer interactions from the analyst. 

Specifically, we claim the following contributions: 
\begin{enumerate}
    \item The $\text{DeepSI}_{\text{finetune}}$ framework that integrates DL into the human-in-the-loop iterative sensemaking pipeline to improve semantic interaction inference.
    \item  
    Two complementary studies, a user-centered qualitative case study and an algorithm-centered simulation-based quantitative experiment, that measure the performance of our method and reveal improvements.
\end{enumerate}

\section{Related Works}
\label{sec:related-works}
Four related components support our design: 
interactive DR models used in semantic interaction;  
basic knowledge of the DL model BERT;  
pretrained DL model adaptation approaches; and  
other work about user-centered interactive DL.

\begin{figure}[htpb]
    \centering
    \includegraphics[width=1.0\columnwidth]{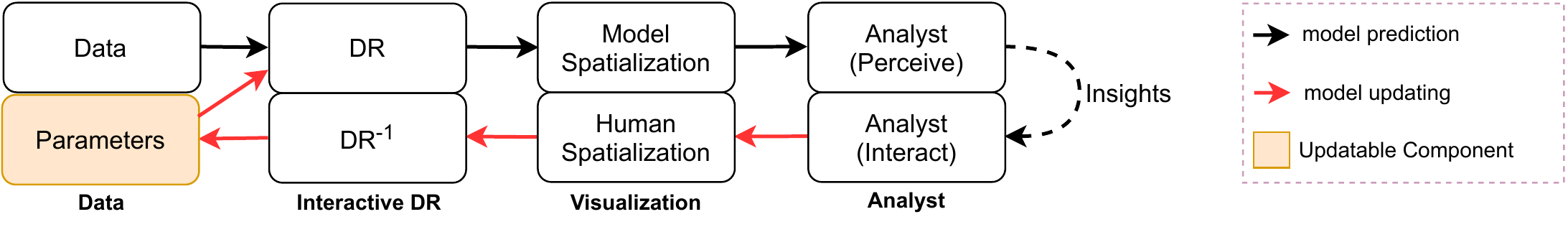}
    \caption{SI pipeline showing the communication between the analyst and VA system, adapted from \cite{Zeitz:2018:BIV:3144687.3144715,endert2012semantic}. 
    The interactive DR component is responsible for capturing the analyst's intent from the human modified projection~(denoted as human spatialization) and, consequently, updating the projection in response~(denoted as model spatialization).}
    \label{fig:bag-of-words-si}
\end{figure}

\subsection{Interactive Dimensionality Reduction}
\label{sec:si}
While using the human-in-the-loop sensemaking SI pipeline~(Fig.~\ref{fig:bag-of-words-si}), analysts gain insights from the model projection~(spatialization) and express their preferences by repositioning data points in the projection.
It is the interactive DR model's responsibility to learn new model parameters that capture the analyst's intent behind the modified projection and, in response, use the learned parameters to update the projection.
Therefore, increasingly powerful DR models have been adapted in a semi-supervised manner to improve SI inference.
VA frameworks V2PI~\cite{Leman:2013it} and BaVA~\cite{House:2015hs} adapted linear DR models, including principal component analysis (PCA)~\cite{WOLD198737} and weighted multidimensional scaling (WMDS)~\cite{schiffman1981introduction}, to the bidirectional SI pipeline.
To support more complex tasks and interactions, multiple models were chained together as a single interactive DR model, which is called multi-model SI~\cite{Bradel:dd,10.1016/j.bdr.2019.04.003,wenskovitch2017observation}. 
Recently, to adapt more powerful but complex non-invertible DR algorithms, such as t-SNE~\cite{maaten2008visualizing} and UMAP~\cite{mcinnes2018umap}, 
Zexplorer~\cite{10.1145/3311790.3396646} used the invertible neural encoder~\cite{espadoto2020deep} to emulate these models as the interactive DR model in SI applications.

Similarly,  $\text{DeepSI}_{\text{finetune}}$ also aims to improve SI inference.
However, $\text{DeepSI}_{\text{finetune}}$ highlights the importance of finding user- and task-specific data representations instead of more powerful DR.
Therefore, we use the simple but commonly used WMDS as the default  DR~\cite{Zeitz:2018:BIV:3144687.3144715,10.1109/VAST.2012.6400486,8440814,self2016bridging} and focus on extracting meaningful DL representations.

\subsection{BERT}
\label{sec:bert}
BERT~(Bidirectional Encoder Representations from Transformers)~\cite{devlin2018bert} is a DL language representation model.
BERT is first pretrained on raw text data to learn general language representations. 
The pretrained BERT then can be easily adapted to downstream NLP tasks such as sentiment analysis and semantic textual similarity~\cite{cer2018universal}. 
The adapted BERT model is able to provide task-specific representations and shows state-of-the-art performance in these downstream tasks.
Technically, BERT is a Transformer encoder~\cite{vaswani2017attention}, containing a stack of transformer layers. 
The transformer layer learns token-level representations. 
For input token sequences, the transformer layer learns a new vector for each token based on all other tokens, using the self-attention mechanism~\cite{bahdanau2014neural}. 
Through the stack of transfer layers, BERT can convert a sequence of tokens into deep representations.
In this paper, we use the pretrained BERT model as the default DL model in the DeepSI pipeline to provide text representations for visual text analytic tasks.

\subsection{Pretrained Representation Adaptation}
There are two main paradigms to adapt the pretrained DL representation model to downstream tasks: feature extraction and fine-tuning~\cite{DBLP:journals/corr/abs-1903-05987}.
The feature extraction approach uses a task-specific architecture to adapt the pretrained representations to downstream tasks (e.g. ELMo~\cite{peters2018deep}).
In this approach, parameters inside the task-specific architecture are trained on the downstream tasks.
In contrast, instead of using a new architecture, the fine-tuning method appends one additional output layer to the pretrained DL and tunes the whole pretrained model with the downstream tasks.
The fine-tuning approach requires relatively less training data because it introduces minimal task-specific parameters and does not need to learn randomly initialized task-specific parameters from scratch. 
Therefore, we use the fine-tuning approach in the $\text{DeepSI}_{\text{finetune}}$ framework to adapt the pre-trained DL model to visual analytic tasks.

\subsection{Human-centered Deep Learning}
There are other human-centered DL techniques proposed to assist users in complex data analytic tasks.
Hsueh-Chien et. al.~\cite{8265023} used CNN techniques to assist users in volume visualization designing through facilitating user interaction with high-dimensional DL features.
In RetainVis~\cite{10.1109/tvcg.2018.2865027}, an interactive and interpretable RNN model was designed for electronic medical records analysis and patient risk predictions, which can be steered interactively by domain experts.
Gehrmann et al.~\cite{10.1109/tvcg.2019.2934595} proposed a framework of collaborative semantic inference that enables the visual collaboration between humans and DL algorithms. 
Sharkzor~\cite{DBLP:journals/corr/abs-1802-05316} is an interactive deep learning system for image sorting and summary, based on users' semantic interactions. 
Of these, Sharkzor is the most similar to our $\text{DeepSI}_{\text{finetune}}$. 
Both works provide users with semantic interactions to tune the DL model interactively. 
However, our work emphasizes a general solution to integrate DL models into SI systems to improve inference.
While Sharkzor is only designed for image analysis, $\text{DeepSI}_{\text{finetune}}$ can be applied to other data analytic tasks and relevant DL models. 

\begin{table}[]
\caption{A list of variables used throughout this paper and their descriptions.}
\label{tab:variables}
\begin{tabular}{p{0.08\textwidth}p{0.85\textwidth}}
    \toprule
    \textbf{\small{Variable}}  & \textbf{\small{Description}} \\
    \toprule
    $d$&\small{A set of documents for analysis}\\
    \midrule
    $N, M$&\small{Number of samples in $d$, number of dimensions of $d$}\\
    \midrule
    $n$&\small{Number of samples moved by the analyst, $n \ll N$}\\
    \midrule
    $x$&\small{High-dimensional feature of $d$.} \footnotesize{DL representations (768 dimension-size BERT embeddings)} \\
    \midrule
    $y$&\small{Coordinates in the 2D visual spatialization of $d$.} \\ 
    \footnotesize{ }&\footnotesize{Set either by analysts' interactions on the visualization, or by the underlying SI model, which maps $x$ to $y$}\\
    \midrule
    $w_{\text{\tiny{dimension}}}$&\small{Parameters of dimension weights of $x$}. \footnotesize{(a 768 dimension-size vector)}\\
    \midrule
    $w_{\text{\tiny{BERT}}}$&\small{Internal parameters of the pretrained BERT model.}  \footnotesize{$\text{BERT}_{base}$ is used in this paper, which contains 110 million parameters}\\
    \midrule
    $dist$&\small{Euclidean distance between data samples in $d$.} \footnotesize{Weighted Euclidean distance is used if $w_{\text{dimension}}$ applied.} \\ 
    \footnotesize{ }&\footnotesize{$dist_{\tiny{H}}$ defines the high-dimensional similarity. $dist_{\tiny{L}}$ defines the low-dimensional similarity
    }\\
  \bottomrule
\end{tabular}
\end{table}

\section{Background}
\label{sec:deepsi-vanilla}

In order to frame our discussion of our model $\text{DeepSI}_{\text{finetune}}$, this section briefly describes DeepVA, the state-of-the-art SI model with pretrained DL~\cite{bian2019deepva}. For the purpose of comparison, we implement a specific version of DeepVA that uses BERT, which we denote as $\text{DeepSI}_{\text{vanilla}}$.
For reference, Table~\ref{tab:variables} describes frequently used variables throughout this paper.
We use the pretrained BERT model as a representative DL model in DeepSI system designs. 
Note that WMDS is used as the default interactive DR in DeepSI frameworks for three reasons. 
First, the WMDS is a simple linear DR algorithm, so that we can focus on assessing the effects of data representations on the model performance. 
Second, WMDS is agnostic to the choice of the weighted distance function.
Third, WMDS enables analysts to express their synthesis process by manipulating data point proximities to reflect their perceived similarity~\cite{Self:2016:BGU:2939502.2939505}. 

\begin{figure*}[htbp]
    \centering
    \includegraphics[width=1.0\textwidth]{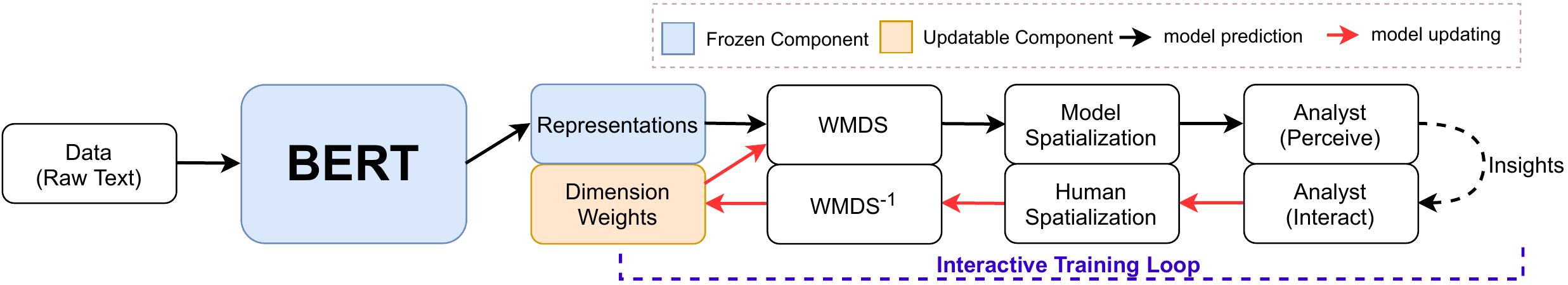}
    \caption{ $\text{DeepSI}_{\text{vanilla}}$ pipeline, adapted from~\cite{bian2019evaluating}: using the pretrained BERT as only a feature extractor pre-processed outside of the interactive loop in SI pipeline.
    All parameters inside the pretrained BERT model are frozen and the output data representations are fixed. 
    WMDS is the interactive DR, which is responsible to tune the dimension weights $\boldsymbol{w}_{\text{\tiny{dimension}}}$ to capture the analyst's intent.
    }
    \label{fig:deepsi-vanilla}
\end{figure*}

$\text{DeepSI}_{\text{vanilla}}$ uses the DL model as only a feature extractor in the SI pipeline. 
As shown in Fig.~\ref{fig:deepsi-vanilla}, the pretrained parameters inside the BERT model are frozen.
Thereby, for an input, BERT provides a fixed general-purpose representation, which is then used as the data features in the interactive training loop.
The BERT model is outside of the interactive loop. 
Therefore, the interactive DR model, WMDS, is responsible for updating dimension weights $\boldsymbol{w}_{\text{\tiny{dimension}}}$ to capture the analyst's intent as a weighting of the BERT features.
The complete process of the pipeline is as follows.

Before entering the interactive training loop, the data representations are initialized by the BERT model with the pretrained parameters $\boldsymbol{w}_{\text{\tiny{BERT}}}$: 
\begin{equation}
    \label{eq:deepsi-vanilla-bert}
    \boldsymbol x = \text{BERT}(d, w_{\text{\tiny{BERT}}})
\end{equation}
In the forward model-prediction direction, WMDS is performed to project high-dimensional data points ($\boldsymbol{x}$) into the two-dimensional spatialization ($\boldsymbol{y}$), with current dimension weights $w_{\text{\tiny{dimension}}}$ (initially, all weights are equal). This provides a new projection for the analyst to perceive and interact.
\begin{equation}
    \label{eq:deepsi-vanilla-wmds}
    \boldsymbol y = \argmin_{y}{\sum_{{\tiny i<j\leq N}}^{} \Big(dist_{\tiny{L}}(y_i, y_j) - dist_{\tiny{H}}(x_i, x_j,  w_{\text{\tiny{dimension}}}) \Big)^2}
\end{equation}
In the backward model-updating direction, the analyst provides visual feedback by repositioning $n$ data points within the projection. 
$\text{WMDS}^{-1}$ uses the low-dimensional pairwise distances between the moved $n$ data points as input, to learn new dimension weights  $\boldsymbol{w}_{\text{\tiny{dimension}}}$ to make sure these moved data points have similar relationships in the high-dimensional space, based on the following optimization criterion:
\begin{equation}
    \label{eq:deepsi-vanilla-mds-1}
    \boldsymbol w_{\text{\tiny{dimension}}} = \argmin_{w}{\sum_{{\tiny i<j\leq n}}^{} \Big(dist_{\tiny{L}}{(y_i, y_j)} - dist_{\tiny{H}}(x_i, x_j, w_{\text{\tiny{dimension}}}) \Big)^2}
\end{equation}
Therefore, through this  loop, the dimension weights $\boldsymbol{w}_{\text{\tiny{dimension}}}$ are trained interactively and incrementally based on analysts' interactions to capture their intents.

$\text{DeepSI}_{\text{vanilla}}$ has been well-evaluated previously.
DeepVA~\cite{bian2019deepva} used ResNet~\cite{he2016deep} as an image data feature extractor in the SI system that assists users performing visual concepts analysis using DL representations. 
In~\cite{bian2019evaluating}, Bian et al. compared SI systems that use embedding vectors as features and those that use bag-of-words as features in visual text analysis tasks.
Experiments in both works show that even the general-purpose representations of pretrained DL models can enable SI to better capture the analyst's intent than hand-crafted features.
However, using the general-purpose pretrained representations still restricts SI inference.
In the next section, we propose $\text{DeepSI}_{\text{finetune}}$, which exploits fine-tuned representations to further improve SI inference.
As the best-performing model from previous studies, $\text{DeepSI}_{\text{vanilla}}$ is the baseline model for comparison.

\section{Model Description}
\label{sec:deepsi-finetune}

\begin{figure*}
    \centering
    \includegraphics[width=1.0\textwidth]{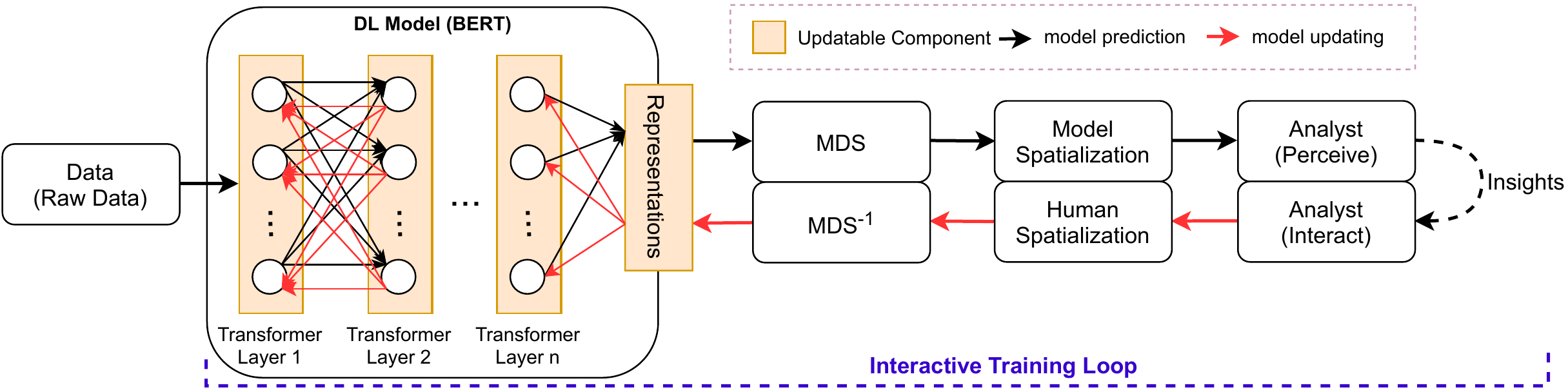}
    \caption{$\text{DeepSI}_{\text{finetune}}$ pipeline: embedding BERT within the SI loop. 
    Semantic interactions are exploited to fine-tune BERT interactively through backpropagation.  
    The tuned BERT is responsible for generating new representations, so as to capture the analyst's intent.
    Thereby, no external parameters are needed.
    }
    \label{fig:deepsi-finetune}
\end{figure*}

This section outlines the main design, model pipeline, and implementation details of  $\text{DeepSI}_{\text{finetune}}$.

\subsection{Model Design}
\label{sec:requirements}
We propose two main design goals to address the two research questions discussed in Sec.~\ref{sec:intro}.

\textbf{Design goal 1 - Integrating DL into the human-in-the-loop interactive sensemaking pipeline.}
To get user- and task-specific representations, it is necessary to iteratively train the DL model with semantic interactions during the human-in-the-loop process.
Inspired by multi-model SI systems~\cite{Bradel:dd,10.1016/j.bdr.2019.04.003,wenskovitch2017observation}, we inserted the DL model update and prediction process into the bidirectional semantic interaction loop, as shown in Fig.~\ref{fig:deepsi-finetune}.
The DL model update and prediction process occurs before the interactive DR model.
The interactive DR model passes the analyst's visual feedback from the human spatialization to the DL model.
The visual feedback is then used to update the parameters inside the DL model ($\boldsymbol{w}_{\text{\tiny{BERT}}}$) through the DL backpropagation~\cite{rumelhart1986learning}~(red arrows inside the DL component).
With updated parameters $\boldsymbol{w}_{\text{\tiny{BERT}}}$, the BERT model calculates new representations for input data by the forward propagation~\cite{zhang2020dive} through the internal transformer layers~(black arrows inside the DL component). 
Through the interactive sensemaking process, the DL model is trained by semantic interactions in an interactive machine learning setting~\cite{amershi2014power,fails2003interactive}.
Thereby, improved representations are generated to accurately capture the analyst's intent. 

\textbf{Design goal 2 - Introducing minimal parameters into the interactive DL training pipeline.}
To solve analytic tasks efficiently, the analyst prefers to perform fewer interactions in each sensemaking loop.
However, DL model training typically needs a relatively large amount of training data. 
To reduce the number of interactions needed for training, 
we should introduce minimal parameters into the pipeline while integrating the DL model training.
For this design goal, we made specific modifications to both the DL and the interactive DR components.
First, we used the fine-tuning approach to adapt the pretrained BERT model with semantic interactions.
Unlike the feature-based method, fine-tuning approach introduced minimal task-specific paramters~\cite{1640964}. 
This drastically reduced the required training data.
Further, we used MDS/$\text{MDS}^{-1}$ as the interactive DR component. 
During the interactive BERT model training, representations are updated to capture analysts' intents. 
It is unnecessary to tune extra parameters for the same purpose in the interactive DR model. 
Therefore, we used MDS/$\text{MDS}^{-1}$ without dimension weights $\boldsymbol{w}_{\text{\tiny{dimension}}}$ as the interactive DR component. 
There are no parameters  to tune in this component. 
Therefore, users' interactions can be passed directly to DL training without  information loss.

\subsection{Model Pipeline}
We illustrate the $\text{DeepSI}_{\text{finetune}}$ pipeline~(Fig.~\ref{fig:deepsi-finetune}) in detail through the human-in-the-loop sensemaking process.
In the forward model-prediction direction, new representations are generated for the dataset $d$ through the forward propagation calculation of the BERT model, with current BERT parameters~($\boldsymbol{w}_{\text{\tiny{BERT}}}$):
\begin{equation}
    \label{eq:deepsi-finetune-bert}
    \boldsymbol x = \text{BERT}(d, w_{\text{\tiny{BERT}}})
\end{equation}
The high-dimensional DL representations $x$ are then projected to the 2D spatialization~(model spatialization) by MDS through the following equation: 
\begin{equation}
    \label{eq:deepsi-finetune-mds}
    \boldsymbol y = \argmin_{y}{\sum_{{\tiny i<j\leq N}}^{} \Big(dist_{\tiny{L}}(y_i, y_j) - dist_{\tiny{H}}(x_i, x_j) \Big)^2}
\end{equation}
In contrast to Eq.~\ref{eq:deepsi-vanilla-wmds}, the high-dimensional distance function is not explicitly weighted. Instead, the updates to $\boldsymbol{y}$  in each loop are captured by the fine-tuned representation $x$ itself. 
The analyst perceives the updated  spatialization and gains insight.

In the backward model-updating direction, the analyst modifies the visual layout~(human spatialization) by repositioning some samples to express the preferred similarities between them.
Then, $\text{MDS}^{-1}$ uses the human-defined similarities between $n$ moved data points, $dist_L(y_i, y_j)$), to steer the BERT model parameters to generate better high-dimensional representations $x$, such that the similarity of the representations reflects the proximity of the points in the modified projection, as follows:  
\begin{equation}
    \label{eq:deepsi-finetune-backpropagation}
   w_{\text{\tiny{BERT}}} = \argmin_{w_{\text{\tiny{BERT}}}}{\sum_{{\tiny i<j\leq n}}^{} \Big(dist_L(y_i, y_j) - dist_H(\text{BERT}(d_i, w_{\text{\tiny{BERT}}}), \text{BERT}(d_j, w_{\text{\tiny{BERT}}})) \Big)^2}
\end{equation}
The optimization objective is to fine-tune BERT weights $\boldsymbol{w}_{\text{\tiny{BERT}}}$ to minimize the difference between low-dimensional and high-dimensional distances of $n$ moved data points through backpropagation. 
All internal parameters of the BERT model~($\boldsymbol{w_{\text{\tiny{BERT}}}}$) are updated in order, from last transformer layers to previous layers, by a gradient descent optimization algorithm~\cite{ruder2016overview}. 
After the backpropagation, the updated $\boldsymbol{w}_{\text{\tiny{BERT}}}$ is used in the forward propagation to calculate new representations $\boldsymbol{x}$ in Eq.~\ref{eq:deepsi-finetune-bert}.

Through this human-in-the-loop interactive DL process, the BERT model is tuned properly to generate user- and task-specific representations so as to capture analysts' precise intents: samples that should be closer to each other in the visualization obtain similar features, while  more distant samples gain  differing features.

\subsection{Prototyping Detail}
Here, we describe the implementation details of $\text{DeepSI}$ prototypes used in our experiments, including model settings and visualization design.
These implementations are applicable for both $\text{DeepSI}_{\text{finetune}}$ and $\text{DeepSI}_{\text{vanilla}}$.

\subsubsection{Model Settings}
We use Pytorch~\cite{NEURIPS2019_9015}, a well-known Python DL framework, to implement the DeepSI system. 
For the forward DR component, MDS is adapted from Scikit-Learn~\cite{scikit-learn}. 
The $\text{MDS}^{-1}$ is implemented in Pytorch as a neural network layer. 
The pretrained BERT model is adapted from the publicly available Python library, Transformers~\cite{Wolf2019HuggingFacesTS}.
Transformers provides two sizes of pretrained BERT models: $\text{BERT}_{\text{BASE}}$, and $\text{BERT}_{\text{LARGE}}$. 
We used the small BERT model~($BERT_{BASE}$)~(bert-base-uncased, 12-layers, 768-hidden, 12-heads, 110M parameters), because it is more stable on small datasets.
For a document containing a list of tokens, $\text{BERT}_{\text{BASE}}$ can convert each of the tokens into a 768-dimensional vector. 
To generate fixed-length encoding vectors from documents of different lengths, we appended a MEAN pooling layer to the last transformer layer of the BERT model, such that the output representation for a document was a 768-dimensional vector.
Therefore, the $w_{\text{\tiny{dimension}}}$ used in $\text{DeepSI}_{\text{vanilla}}$ is also a 768 dimension vector.
We also tested other pooling strategies, such as MAX pooling and CLS pooling~\cite{reimers-2019-sentence-bert}.
However, there was no obvious performance difference, and the MEAN pooling showed slightly better performance.
In addition, we used the Adam optimizer~\cite{kingma2014adam} to optimize the DeepSI model parameters in the model-updating direction. 
We also explored other optimizers provided by PyTorch.
Across all our experiments, we found that Adam optimizer performed the best.
Further, we found that the suggested learning rate~($3e^{-5}$) for finetuning BERT models in~\cite{devlin2018bert} led to optimal $\text{DeepSI}_{\text{finetune}}$ performance in experiments. 

\subsubsection{Visualization Design}
We drew inspiration for visualization design from SI-enabled VA applications, including Andromeda~\cite{Self:2016:BGU:2939502.2939505}, ForSPIRE~\cite{endert2012semantic}, and Dis-Function~\cite{10.1109/VAST.2012.6400486}.
As shown in Fig.~\ref{fig:covid19-1}, the visual interface mainly uses a scatterplot as the projection layout.
This scatterplot not only displays the relationships between data updated by the underlying projection model, but also allows the analyst to intervene and modify the layout.
Specifically, in the forward model-prediction direction, the positions between data points on the scatterplot reflect the points' relative similarity learned by underlying models, either by the projection method or by the fine-tuned BERT model, shown in Fig.~\ref{fig:covid19-1}-1~(model spatialization).
In the backward model-updating direction, the user can drag several data points to new positions to modify similarities between points based on their preference, shown in Fig.~\ref{fig:covid19-1}-2~(human spatialization).
Having both the underlying models and analysts work on the same visualization provides direct and effective communication between humans and computation. 
In addition to the scatterplot view, the prototype also provides a sidebar view to help analysts review the content of a selected document when exploring in the scatterplot view.
In this paper, we intentionally focus on the scatterplot view in the  screenshots, to focus on the analysis of the model performance.

\section{Experiments}
\label{sec:experiment}
To evaluate $\text{DeepSI}_{\text{finetune}}$, we conducted the following experiments. 
To examine how well $\text{DeepSI}_{\text{finetune}}$ addresses the goals, we measured its performance in two respects: 
\begin{itemize}
    \item \textbf{Accuracy}: How accurately can $\text{DeepSI}_{\text{finetune}}$ capture the analyst's intent? 
    \item \textbf{Efficiency}: How many interactions does $\text{DeepSI}_{\text{finetune}}$ need to capture the analyst's intent properly?
\end{itemize}
We use $\text{DeepSI}_{\text{vanilla}}$, described in Sec.~\ref{sec:deepsi-vanilla}, as the baseline model to evaluate the advantage of $\text{DeepSI}_{\text{finetune}}$'s task-specific, instead of general-purpose, representations. 
Boukhelifa et al.~\cite{boukhelifa2018evaluation} proposed the complementary evaluation of interactive machine learning systems by using both algorithm-centered and human-centered evaluation methods.
We perform both evaluation methods in our experiments: the case study in Sec.~\ref{sec:case-study} is the human-centered qualitative analysis, and the simulation-based evaluation method in Sec.~\ref{sec:evaluation} is the algorithm-centered quantitative analysis.

\subsection{Case Study: COVID-19}
\label{sec:case-study}
Recently, COVID-19~\cite{doi:10.1177/0020764020915212} has become a global pandemic. 
It is essential that medical researchers quickly find relevant documents about a specific research question, given the extensive coronavirus literature.
We used an analysis task on academic articles related to COVID-19 in this case study to examine our proposed $\text{DeepSI}_{\text{finetune}}$, compared with the baseline model $\text{DeepSI}_{\text{vanilla}}$.
In this study, we performed the same task with the help of both DeepSI prototypes and then measure the model performance in the following two perspectives:
\begin{itemize}
    \item \textbf{Accuracy}: the quality of the projection updated by the underlying model given the task's ground truth.
    \item \textbf{Efficiency}: how many interactions are needed for the underlying model to provide a useful projection.
\end{itemize}

\subsubsection{Dataset and Task}
\label{sec:covid-19}
The COVID-19 Open Research Dataset~(CORD-19)~\footnote{https://www.kaggle.com/allen-institute-for-ai/CORD-19-research-challenge} contains a collection of more than $200,000$ academic articles about COVID-19. 
CORD-19 also proposes a series of tasks in the form of important research questions about the coronavirus. 
One of the research tasks focuses on identifying COVID-19 risk factors~\footnote{https://www.kaggle.com/allen-institute-for-ai/CORD-19-research-challenge/tasks?taskId=558}. 
In this case study, we selected a task that requires identifying articles related to specific risk factors for COVID-19. 
We asked an expert to choose as many research papers as possible about risk factors from CORD-19.
We found four main risk factors: cancer~(15 articles), chronic kidney disease~(13 articles), neurological disorders~(23 articles), and smoking status~(11 articles). 
We used these four risk factors as the ground truth for the test task, and loaded all these 62 articles into our DeepSI prototypes.
Therefore, the test task was to organize these 62 articles into four clusters with our DeepSI prototype such that each cluster represented articles of a specific risk factor.

This particular ground truth is just one possible way an analyst might want to organize this group of documents. So our goal is to see if this particular set of expert knowledge can be easily injected using SI to help re-organize the documents in this particular way.
To help judge the quality of the visual layout in organizing this particular ground truth, we  color the dots according to the ground-truth risk factors in the visualization:
\textbf{cancer}~(black dot $\textcolor{black}{\bullet}$), 
\textbf{chronic kidney disease}~(red dot $\textcolor{red}{\bullet}$),
\textbf{neurological disorders}~(blue dot  $\textcolor{blue}{\bullet}$),
\textbf{Smoking status}~(green dot $\textcolor{green}{\bullet}$).
It should be noted that the underlying model was not provided with the ground truth or color information. The ground truth is only injected via semantic interaction from the human in the form of partial groupings of only a few of the documents.

\subsubsection{Study Procedure}
\begin{figure*}[htbp]
    \centering
    \includegraphics[width=1.0\textwidth]{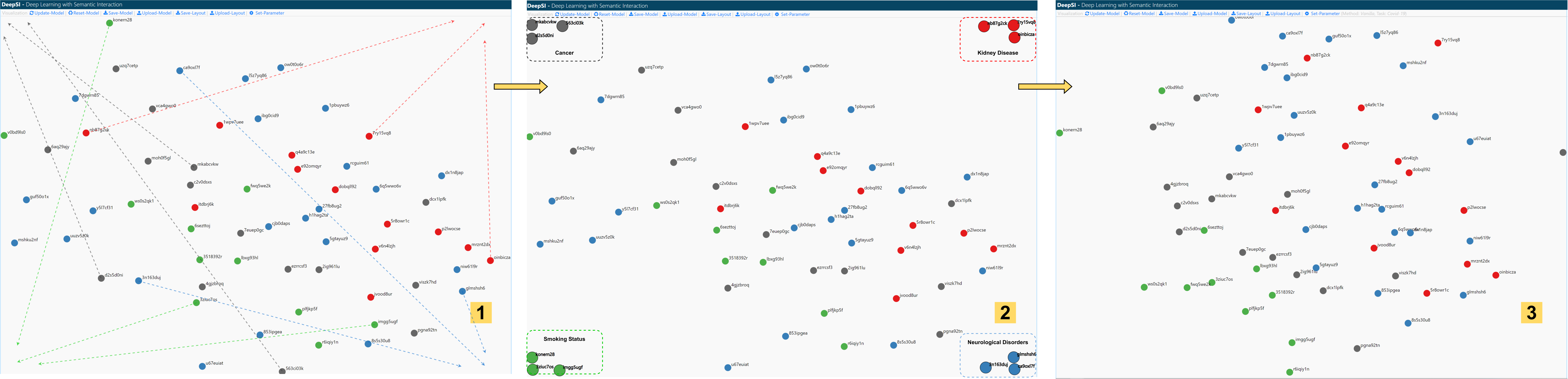}
    \caption{Screenshots during the case study using $\text{DeepSI}_{\text{vanilla}}$: 
    Frame 1 and 2 show the similar initial steps performed by the analyst in~Fig.~\ref{fig:covid19-1}.
    Frame 3 shows the resulting projection updated by $\text{DeepSI}_{\text{vanilla}}$.
    }
    \label{fig:covid19-2}
\end{figure*}

To compare the projection layouts updated by two models based on the same input interactions, semantic interactions based on the ground truth are performed in the shared visual projection and then applied to the two models separately.
Fig.~\ref{fig:covid19-1} and Fig.~\ref{fig:covid19-2} show the process of interactions applied separately to $\text{DeepSI}_{\text{finetune}}$ and $\text{DeepSI}_{\text{vanilla}}$ prototypes.
In both figures, frame 1 and frame 2 are from the shared visual projection.
Frame 1 in both figures~(Fig.~\ref{fig:covid19-1}-1 and Fig.~\ref{fig:covid19-2}-1) shows the same initial layout updated by the default pretrained BERT model.
In the initial projection layout, all the articles are combined.
This means that the pretrained BERT model cannot distinguish these articles by their related risk factors.
Interactions were performed within the projection based on the ground truth to reflect the perceived connections between articles:
grouping three articles about cancer to the top-left region of the projection, indicated by the black arrows;
three articles about chronic kidney disease to the top-right region indicated by red arrows;
three articles about smoking status to the bottom-left part indicated by green arrows;
and three articles about about neurological disorders to the bottom-right part indicated by blue arrows.

Frame 2~(Fig.~\ref{fig:covid19-1}-2 and Fig.~\ref{fig:covid19-2}-2) is the same human spatialization and shows four clusters created by the ground truth.
After we clicked the `model update' button on the menu bar to start the model training process.
Then, the same human spatialization was used to train both models.
After the two models had been updated, the updated projections of these two models showed the performance difference in frame 3 (Fig.~\ref{fig:covid19-1}-3 and Fig.~\ref{fig:covid19-2}-3).
Subsequently, the performance of the model could be assessed based on how reasonable the layout was in comparison with the ground truth. 

\textbf{$\text{DeepSI}_{\text{finetune}}$ spatialization:} 
The projection updated by $\text{DeepSI}_{\text{finetune}}$ is shown in Fig.~\ref{fig:covid19-1}-3.
There are four clear clusters, and all articles are clearly grouped into the correct clusters. 
The top left cluster contains all the articles about cancer~($\textcolor{black}{\bullet}$), 
the top right cluster contains articles about kidney disease~($\textcolor{red}{\bullet}$),
the bottom left contains articles about smoking status~($\textcolor{green}{\bullet}$),
and the bottom right contains articles about neurological disorders~($\textcolor{blue}{\bullet}$).
This means the new representations generated by the fine-tuned BERT model are able to accurately capture the semantic meanings behind users' interactions. 

\textbf{$\text{DeepSI}_{\text{vanilla}}$ spatialization:} 
With the same interactions as input, the updated $\text{DeepSI}_{\text{vanilla}}$ shows a different layout.
As shown in Fig.~\ref{fig:covid19-2}-3, there are no clear clusters in the updated layout compared with Fig.~\ref{fig:covid19-1}-3.
Articles about different risk factors still overlap.
Even after continued interactions based on the ground truth, $\text{DeepSI}_{\text{vanilla}}$ is unable to properly capture the user's semantic intent and differentiate these articles.

\begin{figure*}
    \centering
    \includegraphics[width=1.0\textwidth]{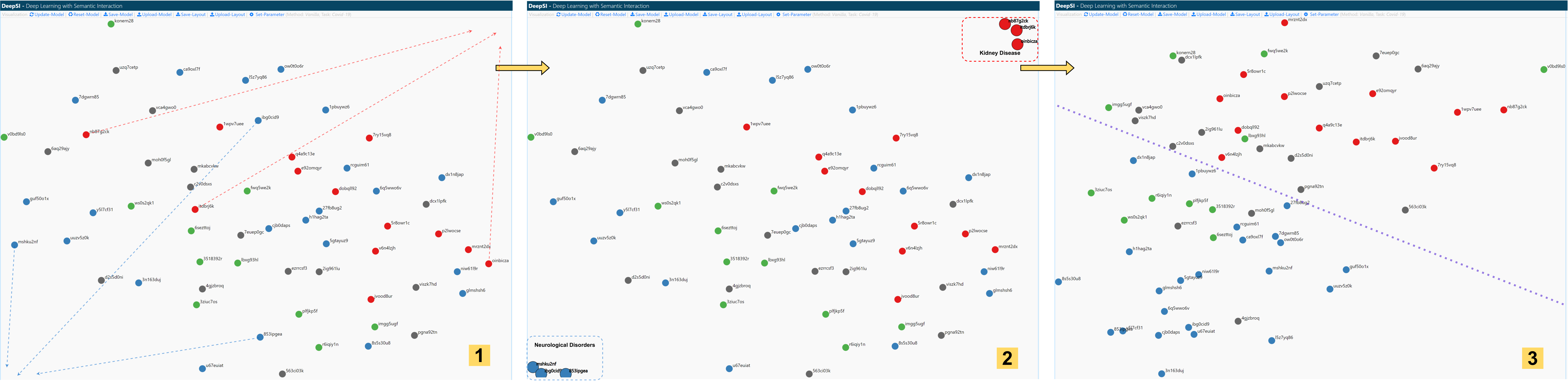}
    \caption{Further case study using $\text{DeepSI}_{\text{vanilla}}$ in grouping two clusters: 
    Frame 1 is the initial projection layout,
    Frame 2 shows interactions performed within the projection, 
    and Frame 3 shows the resulting projection updated by $\text{DeepSI}_{\text{vanilla}}$.
    }
    \label{fig:covid19-3}
\end{figure*}

\textbf{Further study for $\text{DeepSI}_{\text{vanilla}}$:} 
However, $\text{DeepSI}_{\text{vanilla}}$ did work well at separating articles into two clusters in two opposite positions in the projection.
For example, in Fig~\ref{fig:covid19-2}-3, smoking status articles ($\textcolor{green}{\bullet}$) are separated from kidney disease articles ($\textcolor{red}{\bullet}$).
Exploring further, after resetting the model 
as shown in Fig.~\ref{fig:covid19-3}-2,  three neurological disorders articles~($\textcolor{blue}{\bullet}$) are dragged to the bottom-left and three chronic kidney disease articles~($\textcolor{red}{\bullet}$) are dragged to the top-right on the scatterplot view.
After the layout updates, articles from these two dragged clusters are well placed in two opposite sides of the visualization in Fig.~\ref{fig:covid19-3}-3, ignoring articles in the other two clusters~(about cancer $\textcolor{black}{\bullet}$ and smoking status $\textcolor{green}{\bullet}$).

\subsubsection{Qualitative Results}
In terms of accuracy, $\text{DeepSI}_{\text{finetune}}$  grouped articles  correctly based on the user-defined risk factors. 
In contrast, $\text{DeepSI}_{\text{vanilla}}$ did not provide a useful projection.
The further study also confirmed that $\text{DeepSI}_{\text{vanilla}}$ can  handle  more straightforward tasks with only two clusters. 
The semantics of the ground truth knowledge provided by the contest organizers are more recognizable in the  $\text{DeepSI}_{\text{finetune}}$ projection.
Articles in each group are clearly clustered.
However, $\text{DeepSI}_{\text{vanilla}}$  only partially separated in two separate directions, instead of into distinct clusters, which requires more cognitive effort to identify the boundary between the groups.
In terms of efficiency, $\text{DeepSI}_{\text{finetune}}$ is more efficient than $\text{DeepSI}_{\text{vanilla}}$.
$\text{DeepSI}_{\text{finetune}}$ needed a  small number of interactions~(moving three articles in each cluster, 12 dots movement in total)~in one interactive SI loop to fine-tune the BERT model properly for this task.
In contrast, in $\text{DeepSI}_{\text{vanilla}}$, the same amount of interactions  only supported the simpler task with two clusters, and additional rounds of interaction still did not uncover all four clusters.

\begin{figure*}
  \centering
  \includegraphics[width=\textwidth]{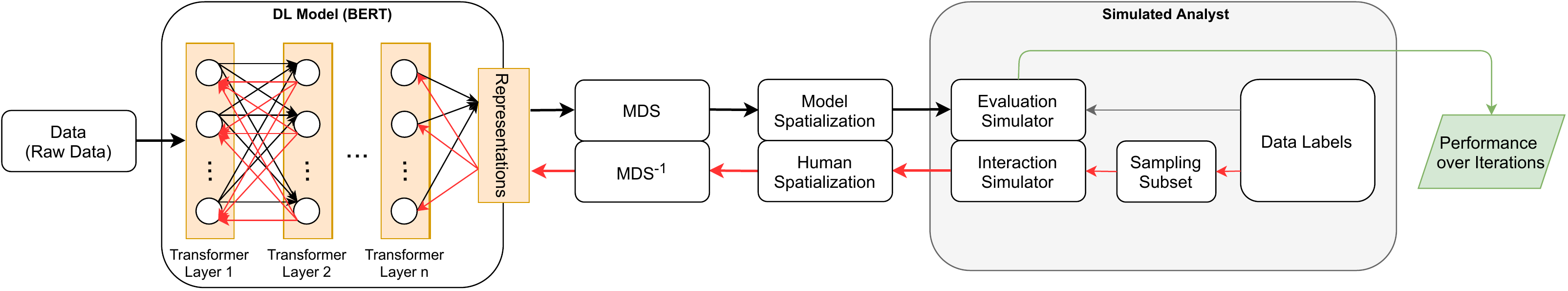}
  \caption{
  Simulation-based evaluation pipeline.
  The analyst is replaced by the `simulated analyst' component where:
  analyst perception is simulated by the kNN classifier and analyst interaction by sampling a subset of ground truth.
  In each SI loop, the kNN classification is employed to calculate the accuracy of the model spatialization, which is updated by the underlying model $\text{DeepSI}_{\text{finetune}}$ and reflects the model performance.}
  \label{fig:simulation-pipeline}
\end{figure*}

\subsection{Simulation-based Evaluation}
\label{sec:evaluation}
From the machine learning algorithm perspective, DeepSI systems are transductive models~\cite{6813505} that interactively learn projections provided by the analyst.
Therefore, the performance of DeepSI systems can be measured by the predicted projections.
To conduct the quantitative comparisons between the predicted projections from DeepSI systems, we replaced the analyst with a simulation component~(simulated analyst).
As shown in Fig.~\ref{fig:simulation-pipeline}, the simulated analyst uses the interaction simulator to generate a training projection~(human spatialization) based on data labels, and the evaluation simulator to evaluate the accuracy of the predicted projection.
After training iteration, the simulated analyst outputs a current projection accuracy.
The projection accuracy over iterations reflects the learning curve~\cite{10.1162/153244304322972694} of the DeepSI model.
Therefore, performances of both DeepSI models could be compared through their learning curves in both accuracy and efficiency perspectives.

\subsubsection{Simulated Analyst}
As shown in the simulation pipeline~(Fig.~\ref{fig:simulation-pipeline}), data labels are the ground truth to support both interaction simulator and evaluation simulator.
First, the interaction simulator uses these labels to calculate the pairwise distances between a subset of data samples, simulating the human-defined similarities between these samples.
Further, the evaluation simulator uses these class labels to measure how well the predicted projection  grouped data samples into correct classes based on their labels.

\textbf{Interaction simulator:}
In each interaction, three samples from each class are   selected using random sampling~\cite{tille2006sampling}. 
Then the interaction simulator calculates the pairwise distance $dist_L{(y_i, y_j)}$ of these selected samples based on:
\begin{equation*}
dist_L(d_i, d_j) = 
\begin{cases}
0 & \text{if $d_i$ and $d_j$ have the same label}\\
\sqrt{2} & \text{otherwise}
\end{cases}
\label{eq:simulation}
\end{equation*}
As shown in the above Equation, if two selected samples have different labels, the distance between them is~$dist_L(d_i, d_j) = \sqrt{2}$, because the analyst should move them away from each other to obtain the farthest distance on the 2D spatialization.
If the two samples have the same label, the analyst should move them as close as possible~($dist_L(d_i, d_j) = 0$) in the projection, because they belong to the same cluster.
Therefore, the interaction simulator provides the calculated pairwise distances between the selected samples as the training projection for DeepSI models.

\textbf{Evaluation simulator:}
After the interaction simulator trains the DeepSI model, the trained model predicts a new projection.
The predicted projection reflects the similarity relationships between samples in the low-dimensional spatialization.
We used a kNN~(K-nearest-neighbour) classifier~\cite{Cover:2006:NNP:2263261.2267456}~ as the evaluation simulator to measure the predicted projection~\cite{10.1109/VAST.2012.6400486,bian2019evaluating}. 
The kNN classifier uses the neighbor information on the projection to train and predict the data classes.
The performance of the learned kNN classifier can directly reflect the quality of the projection~\cite{weinberger2009distance}.
Concretely, we used the leave-one-out cross-validation~\cite{10.5555/645529.658292} and set k = 5 closest training examples to predict the unlabelled sample. We also explored other values for k, such as 3, 7, 9, 11, but these did not produce significant changes in the results.
We could thus obtain the trained kNN classifier accuracy by comparing the predicted output with the ground truth. 

\textbf{Performance over iteration:}
A new accuracy from the  kNN classifier was returned from the simulation pipeline in each iteration loop. These are accumulated into 
 a plot of kNN classifier performance over the iterations of the simulated interaction loop.
This learning curve shows how rapidly the DeepSI model  learned during the interactive  process.

\subsubsection{Dataset and Task}
We explored three commonly used text corpora in natural language processing and visual text analysis tasks. 
These corpora contain different numbers of labels and are from different domains, providing a comprehensive evaluation of performance comparisons.  

\textbf{SST with two clusters:}
The SST dataset~\cite{socher2013recursive} is a collection of movie reviews with both fine-grained labels~(out of five stars) and binary labels~(positive and negative reviews).
We used the binary version of the dataset, which contains 1821 reviews in total: 909 positive and 912 negative.
The task, denoted as $T_{\text{sst}}$, used the SST dataset to train the DeepSI methods to obtain two clusters~(positive and negative).

\textbf{Vispubdata with three clusters:}
The Vispubdata dataset~\cite{Isenberg:2017:VMC} contains academic papers published in the IEEE VIS conference series. 
These papers belong to one of the three conferences: InfoVis~(Information Visualization), SciVis~(Scientific Visualization), and VAST~(Visual Analytics Science and Technology).
We used the papers published between 2008 and 2018~(including 397 papers from InfoVis, 534 papers from SciVis, and 521 papers from VAST) in this task, denoted as $T_{\text{vis}}$.
In $T_{\text{vis}}$, the simulated analyst need to iteratively drag papers into these three conferences clusters to evaluate the DeepSI.

\textbf{20 Newsgroups with four clusters:}
The 20 Newsgroup dataset~\footnote{http://qwone.com/~jason/20Newsgroups/} is a collection of newsgroup posts on 20 topics. 
Based on this dataset, we create the task~($T_{\text{news}}$) to classify four topics into different clusters in the spatialization.
We picked four topics from the same sub-category `rec' including: 594 reports from `rec.autos', 598 reports from `rec.motorcycles', 597 reports from `rec.sport.baseball', and 600 reports from `res.sport.hockey'.

\begin{figure}[h]
    \centering
    \subfigure[$T_{\text{sst}}$]{\includegraphics[width=0.325\textwidth]{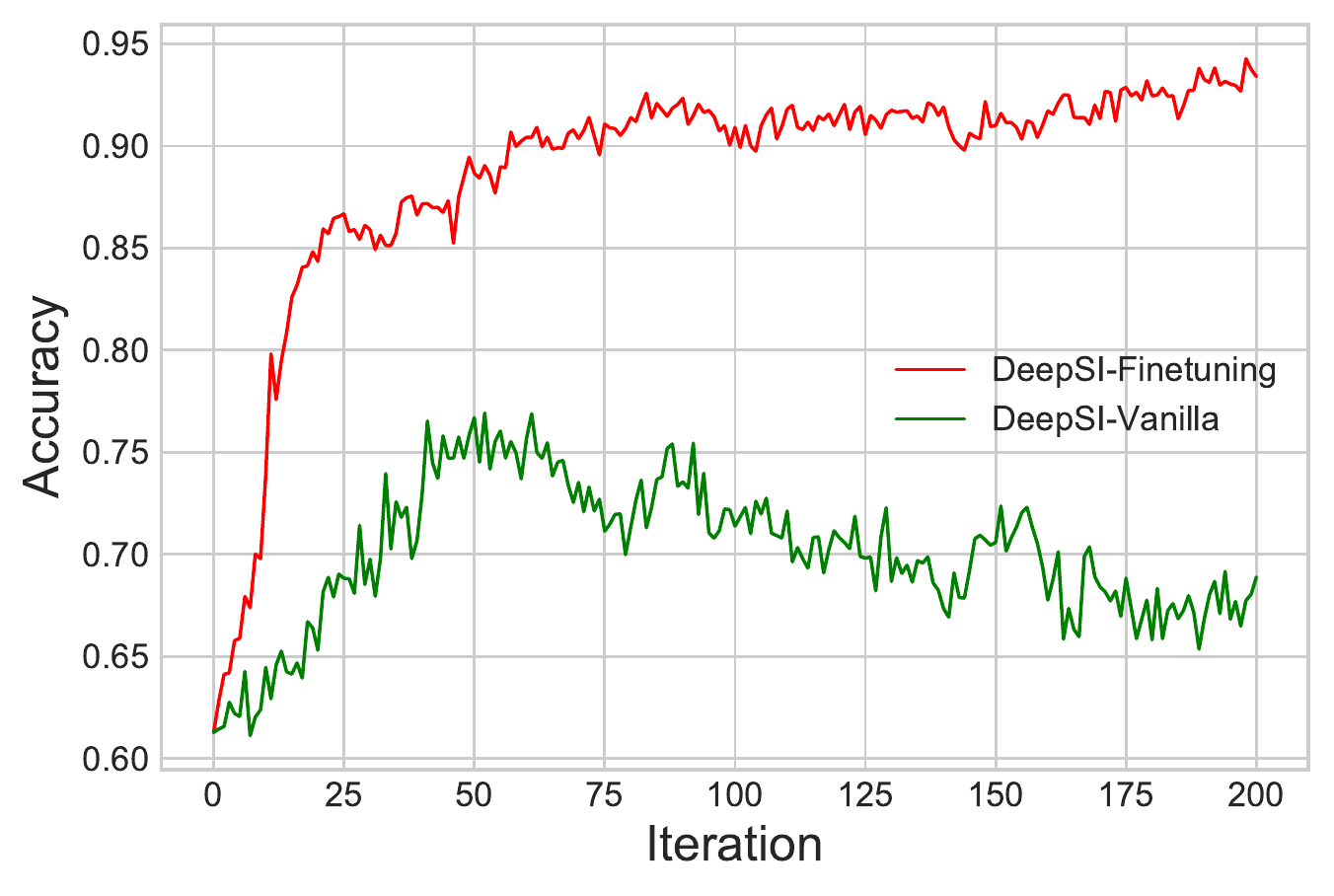}
    \label{fig:task-sst}
    } 
    \subfigure[$T_{\text{vis}}$]{\includegraphics[width=0.325\textwidth]{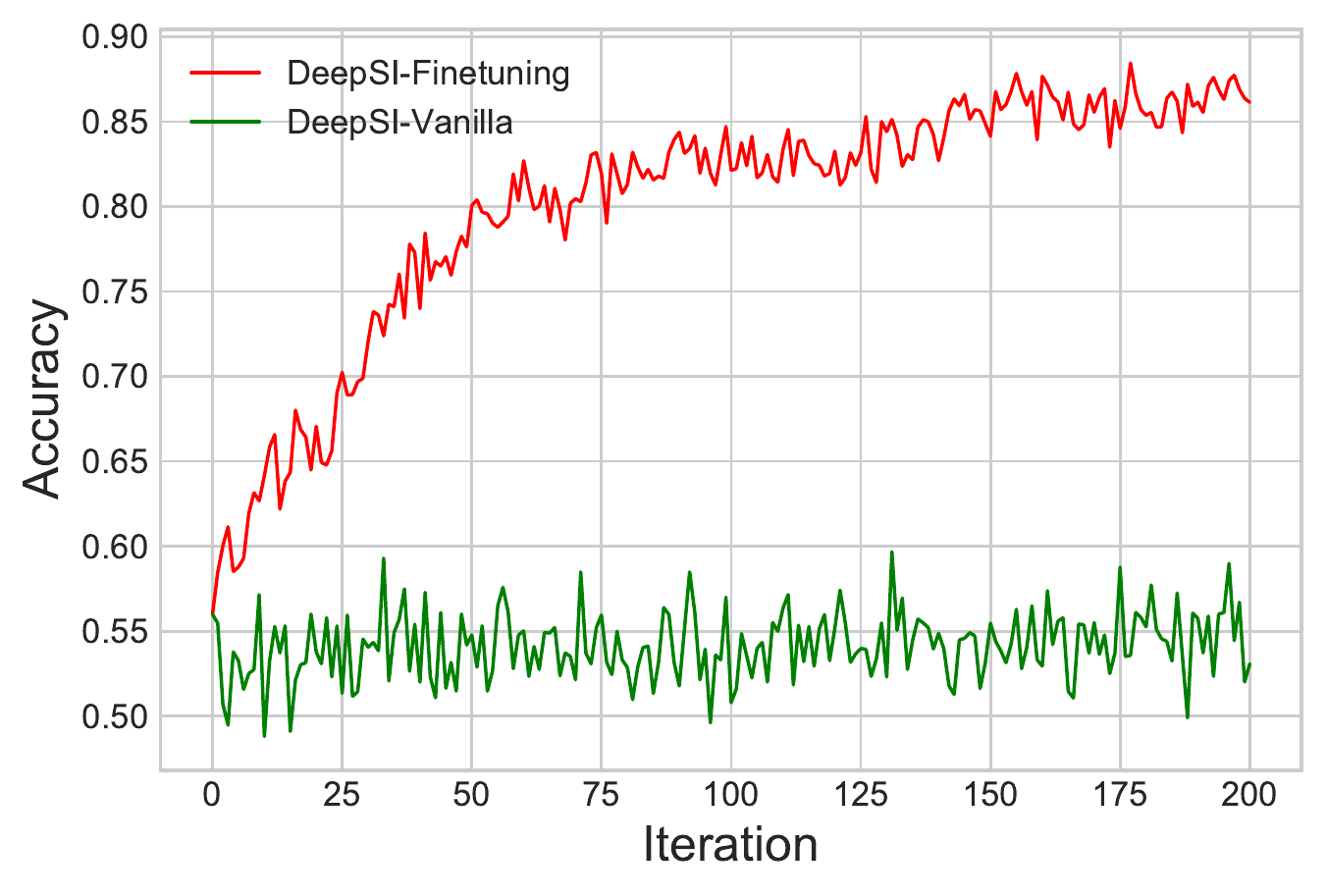}
    \label{fig:task-vis}
    } 
    \subfigure[$T_{\text{news}}$]{\includegraphics[width=0.325\textwidth]{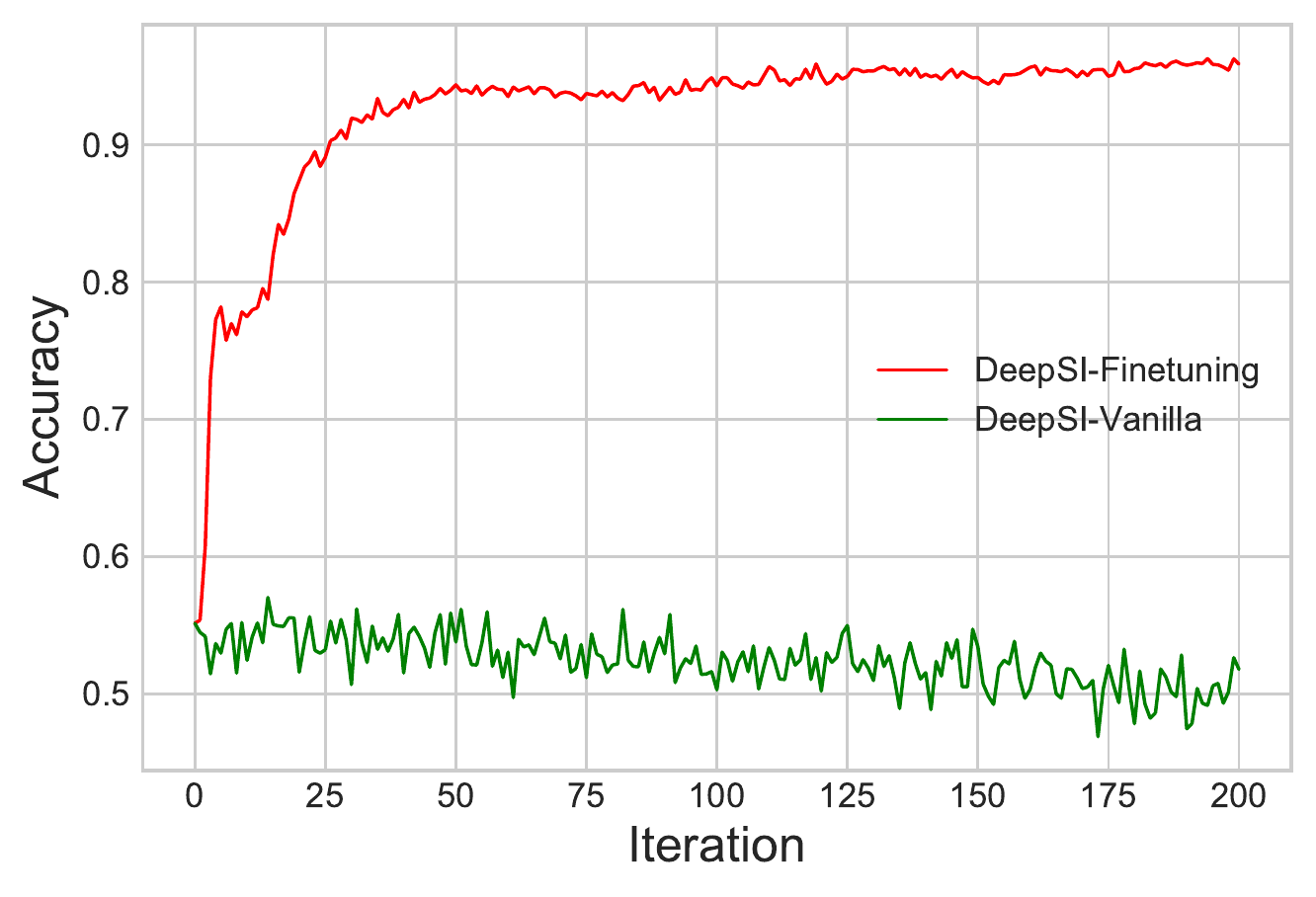}
    \label{fig:task-news}
    }
    \caption{The accuracies of both $\text{DeepSI}_{\text{finetune}}$ and $\text{DeepSI}_{\text{vanilla}}$ updated projections over 200 iterations across the three tasks~($T_{\text{sst}}$, $T_{\text{vis}}$, and $T_{\text{news}}$) during the simulation-based experiment.}
    \label{fig:simulation-result}
\end{figure}

\subsubsection{Quantitative Results}
Fig.~\ref{fig:simulation-result} shows the learning curves of both DeepSI methods in all three tasks.
There is no crossing of the curves between these two models, and the performance curve of $\text{DeepSI}_{\text{finetune}}$ is above the curve of $\text{DeepSI}_{\text{vanilla}}$ through all iterations. 
This means $\text{DeepSI}_{\text{finetune}}$ showed better performance on all three tasks than $\text{DeepSI}_{\text{vanilla}}$.
For model accuracy, $\text{DeepSI}_{\text{finetune}}$ converged to more than or nearly $90\%$ accuracy in all three tasks.
On the contrary, the best performance of $\text{DeepSI}_{\text{vanilla}}$ only showed slightly higher accuracy~(less than $80\%$) than the initial performance in the first task~($T_{\text{sst}}$) with two clusters.
For tasks with more than two clusters~($T_{\text{vis}}$ in Fig.~\ref{fig:task-vis} and $T_{\text{news}}$ in Fig.~\ref{fig:task-news}), $\text{DeepSI}_{\text{vanilla}}$ did not show noticeable accuracy increases.
This is consistent with our findings in the case study~(Sec.~\ref{sec:case-study}).
For model efficiency, the performance over iterations of $\text{DeepSI}_{\text{finetune}}$ in all three tasks showed steeper increases and quickly approximated peak accuracy.
Furthermore, the performance of $\text{DeepSI}_{\text{finetune}}$ increased fairly consistently compared to $\text{DeepSI}_{\text{vanilla}}$.
This provides analysts with more consistent feedback over iterations.

\section{Discussion}

\subsection{Generality and Applicability}
In this paper, we use  pretrained BERT as the specific DL model in $\text{DeepSI}_{\text{finetune}}$ to advance SI-enabled applications.
Considering $\text{DeepSI}_{\text{finetune}}$ as a framework, it is general enough to apply other pretrained DL models into the semantic interaction pipeline for other VA tasks.
First, other transformer-based models, such as RoBERTa~\cite{liu2019roberta}, XLNet~\cite{yang2020xlnet} and GPT-3~\cite{brown2020language}, can be applied directly in the DeepSI pipeline without any special configuration.
In addition, fine-tunable DL models with other structures, such as CNN and RNN models~\cite{ABIODUN2018e00938}, can also be integrated into a DeepSI pipeline by appending a special pooling layer to transform hidden states into proper representations. 
Further, other feature-based DL models could also be applied to the inteactive fine-tune process with specific designs. 
For example, ELMo~\cite{peters2018deep} could be fine-tuned by using max-pool over the model's internal states and adding a softmax layer~\cite{DBLP:journals/corr/abs-1903-05987}.

\subsection{Scalability}
Beyond measuring accuracy and efficiency, our two experiments also illuminated  scalability. 
All our experiments were conducted on a desktop computer with an Intel i9-9900k processor, 32G Ram, and one NVIDIA GeForce RTX 2080Ti GPU, running Windows 10.
In the case study, $\text{DeepSI}_{\text{finetune}}$  captured the analyst's intent and provided an  accurate projection with 62 data points in real time. 
In the simulation-based experiment, $\text{DeepSI}_{\text{finetune}}$  also provided accurate projections that contains thousands of data points.
During the simulation, MDS projection calculation ($O(n^2)$ algorithm) consumed the majority of the time. 
The amount of time required for the DL model update and prediction was negligible in comparison.
This highlights the potential for improving the DR method in DeepSI.

\subsection{Interactive Deep Metric Learning}
\label{sec:metric-learning}
Dis-Function~\cite{10.1109/VAST.2012.6400486} describes the WMDS-based SI model as an interactive distance function learning model from the interactive machine learning perspective.
Likewise, $\text{DeepSI}_{\text{finetune}}$ can be regarded as an interactive deep metric learning model~\cite{10.3390/sym11091066}.
As shown in Fig.~\ref{fig:deepsi-finetune}, in the model-updating direction, the underlying pretrained BERT model is trained interactively to output better presentations that can capture the analyst-desired distance relationships.
Deep metric learning methods usually use metric loss functions~\cite{10.3390/sym11091066} for labelled data, such as contrastive loss~\cite{1640964}, triplet loss~\cite{10.1007/978-3-319-24261-3_7} and angular loss~\cite{Wang_2017_ICCV}.
In contrast, $\text{DeepSI}_{\text{finetune}}$ uses a metric loss function specially designed for semantic interactions based on $\text{MDS}^{-1}$. 


\subsection{Limitations and Future Work}
Our $\text{DeepSI}_{\text{finetune}}$ proved effective in capturing analysts' precise intents and displaying intuitive projections.
However, current $\text{DeepSI}_{\text{finetune}}$ prototypes could be extended in two directions. 
First, it is important to make the internal status of the underlying model interpretable to analysts in order to facilitate them making hypotheses and decisions during the sensemaking process, which is known as interpretable machine learning~\cite{murdoch2019interpretable}.
Other than the projection scatterplot, the current $\text{DeepSI}_{\text{finetune}}$ prototype does not  provide any other visual hints about the status of the DL model.
Therefore, we plan to add more specific visual designs in future work to better expose the effects of tuning the DL model.

In addition, $\text{DeepSI}_{\text{finetune}}$ uses the traditional metric learning method~\cite{doi:10.2200/S00626ED1V01Y201501AIM030} as the interactive DR component to communicate between the DL model and the analyst. 
As discussed above~(Sec.~\ref{sec:metric-learning}), a MDS-based loss function $\text{Loss}_{\text{\Tiny{SI}}}$ is used to interactively tune the DL model. 
Inversely, these deep metric learning loss functions, such as contrastive loss and triplet loss, could also be used as interactive DR components. 
We plan in future work to use deep metric learning loss functions as the interactive DR component in $\text{DeepSI}_{\text{finetune}}$. 

\section{Conclusion}
In this work, we focused on DeepSI and the research question of how to integrate the DL model into the SI pipeline to leverage its capability to better capture the semantics behind user interactions.
We identified two design requirements of effective DeepSI systems: the DL model is trained interactively in the SI pipeline and the DL model can be tuned properly with a small number of interactions.
We presented $\text{DeepSI}_{\text{finetune}}$, which incorporates DL fine-tuning  and MDS-based interactive DR methods into the DeepSI pipeline to meet these requirements.
We performed two complementary experiments to measure the effectiveness of $\text{DeepSI}_{\text{finetune}}$, including a case study of a real-world task relating to COVID-19 and a simulation-based quantitative evaluation method on three commonly used text corpora.
The results of these two experiments demonstrated that $\text{DeepSI}_{\text{finetune}}$ improves performance over the state-of-the-art alternative that uses DL only as a pre-processed feature extractor, indicating the importance of integrating the DL into the interactive loop.
With a small number of semantic interactions as input, $\text{DeepSI}_{\text{finetune}}$ better captures the semantic intent of the analyst behind these interactions.

\begin{acks}
This work was supported in part by NSF I/UCRC CNS-1822080 via the NSF Center for Space, High-performance, and Resilient Computing (SHREC).
\end{acks}

\bibliographystyle{ACM-Reference-Format}
\bibliography{sample-base}



\end{document}